\pdfoutput=1
\documentclass[letterpaper, paper,11pt]{AAS}	

\usepackage{bm}
\usepackage{amsmath}
\usepackage{subfigure}
\usepackage[colorlinks=true, pdfstartview=FitV, linkcolor=black, citecolor= black, urlcolor= black]{hyperref}
\usepackage{subfiles}
\usepackage{overcite}
\usepackage{footnpag}	
\usepackage{booktabs}
\usepackage{makecell}

\newcommand{\xxnote}[3]{}
\ifx\hidenotes\undefined
  \usepackage{color}
  \renewcommand{\xxnote}[3]{\color{#2}{#1: #3}}
\fi

\PaperNumber{19-840}

\begin{document}

\title{TOWARDS ROBUST LEARNING-BASED POSE ESTIMATION OF NONCOOPERATIVE SPACECRAFT}

\author{Tae Ha Park\thanks{Ph.D.~Candidate, Department of Aeronautics \& Astronautics, Stanford University, Stanford, CA 94305}, Sumant Sharma\footnotemark[1], Simone D'Amico\thanks{Assistant Professor, Department of Aeronautics \& Astronautics, Stanford University, Stanford, CA 94305}
}

\maketitle{} 	

\begin{abstract}
This work presents a novel Convolutional Neural Network (CNN) architecture and a training procedure to enable robust and accurate pose estimation of a noncooperative spacecraft. First, a new CNN architecture is introduced that has scored a fourth place in the recent Pose Estimation Challenge hosted by Stanford's Space Rendezvous Laboratory (SLAB) and the Advanced Concepts Team (ACT) of the European Space Agency (ESA). The proposed architecture first detects the object by regressing a 2D bounding box, then a separate network regresses the 2D locations of the known surface keypoints from an image of the target cropped around the detected Region-of-Interest (RoI). In a single-image pose estimation problem, the extracted 2D keypoints can be used in conjunction with corresponding 3D model coordinates to compute relative pose via the Perspective-n-Point (PnP) problem. These keypoint locations have known correspondences to those in the 3D model, since the CNN is trained to predict the corners in a pre-defined order, allowing for bypassing the computationally expensive feature matching processes. The proposed architecture also has significantly fewer parameters than conventional deep networks, allowing real-time inference on a desktop CPU. This work also introduces and explores the texture randomization to train a CNN for spaceborne applications. Specifically, Neural Style Transfer (NST) is applied to randomize the texture of the spacecraft in synthetically rendered images. It is shown that using the texture-randomized images of spacecraft for training improves the network's performance on spaceborne images without exposure to them during training. It is also shown that when using the texture-randomized spacecraft images during training, regressing 3D bounding box corners leads to better performance on spaceborne images than regressing surface keypoints, as NST inevitably distorts the spacecraft's geometric features to which the surface keypoints have closer relation.
\end{abstract}


\section{Introduction}

The ability to accurately determine and track the pose (i.e., the relative position and attitude) of a noncooperative client spacecraft with minimal hardware is an enabling technology for current and future on-orbit servicing and debris removal missions, such as the RemoveDEBRIS mission by Surrey Space Centre\cite{removedebris}, the Phoenix program by DARPA\cite{phoenix_darpa}, the Restore-L mission by NASA\cite{restore_L}, and GEO servicing programs proposed by Infinite Orbits\footnote[3]{www.infiniteorbits.io}, Effective Space\footnote[4]{https://www.effective.space/}, and many other startup companies. In particular, performing on-board pose estimation is key to the real-time generation of the approach trajectory and control update. The use of a single monocular camera to perform pose estimation is especially attractive due to the low power and mass requirements posed by small spacecraft such as CubeSats. Previous approaches to monocular-based pose estimation\cite{damico_benn_jorgensen_2014, Sharma2018, Capuano2018} employ image processing techniques to detect relevant features from a 2D image, which are then matched with features of a known 3D model of the client spacecraft in order to extract relative attitude and position information\cite{Sharma2016}. However, these approaches are known to suffer from a lack of robustness due to low signal-to-ratio, extreme illumination conditions, and dynamic Earth background in space imagery. Moreover, these approaches are computationally demanding during pose initialization due to a large search space in determining the feature correspondences between the 2D image and the 3D model.

On the other hand, recent advances in terrestrial computer vision applications of object pose estimation incorporate deep learning algorithms\cite{Tulsiani2015, Su2015, Poirson2016, Kehl2017, Sundermeyer_2018_ECCV, Kendall2015, Mahendran2017, Xiang2018, Li2018, Rad2017, Tekin2018, Tremblay2018, Zhao2018Estimating6P, Peng2019_PVNet}. Instead of relying on explicit, hand-engineered features to compute the relative pose, these algorithms based on deep Convolutional Neural Networks (CNN) are trained to learn the nonlinear mapping between the input images and the output labels, often six-dimensional (6D) pose space or other intermediate information to compute the relative pose. For example, PoseCNN \cite{Xiang2018} directly regresses relative attitude expressed as a unit quaternion and relative position via separate CNN branches, whereas the network of Tekin et al.\cite{Tekin2018}~modifies the YOLOv2 object detection network \cite{redmon_farhadi_2017} to regress the 2D corner locations of the 3D bounding box around the object. The detected 2D corner locations can be used in conjunction with corresponding 3D model coordinates to solve the Perspective-n-Point (PnP) problem \cite{Lepetit2008} and extract the full 6D pose. Similarly, KeyPoint Detector (KPD) \cite{Zhao2018Estimating6P} first uses YOLOv3 \cite{redmon_farhadi_2018} to localize the objects, then uses ResNet-101 \cite{He2015_ResNet} to predict the locations of SIFT features \cite{Lowe2004} which can be used in the PnP problem to compute the 6D pose. Recently, PVNet \cite{Peng2019_PVNet} architecture is proposed to regress the pixel-wise unit vectors which are then used to vote for the location of the keypoints in a way similar to the Random Sample Consensus (RANSAC) \cite{Fischler1987} algorithm. The RANSAC-based voting scheme allows improved prediction accuracy on occluded and truncated objects. PVNet achieves significantly improved performance on LINEMOD and OccludedLINEMOD benchmark datasets \cite{Hinterstoisser2013_LINEMOD, Brachmann2014}.

Not surprisingly, several authors have recently proposed to apply deep CNN to spaceborne pose estimation\cite{SharmaBeierle2018, Shi2018_CubesatCNN, Sharma2019}. Notably, the recent work of Sharma and D'Amico introduced a CNN-based Spacecraft Pose Network (SPN) with three branches that solves for the pose using state-of-the-art object detection network and the Gauss-Newton algorithms\cite{Sharma2019}. The same work also introduced the Spacecraft Pose Estimation Dataset (SPEED) benchmark that contains 16,000 images consisting of synthetic and real camera images of a mock-up of the Tango spacecraft from the PRISMA mission\cite{damico_benn_jorgensen_2014, PRISMA_chapter}. The dataset is publicly available for researchers to evaluate and compare the performances of pose estimation algorithms and neural networks. Moreover, the SPEED was used in the recent Satellite Pose Estimation Challenge\footnote{https://kelvins.esa.int/satellite-pose-estimation-challenge/} organized by the Stanford University's Space Rendezvous Laboratory (SLAB) and the Advanced Concepts Team (ACT) of the European Space Agency (ESA).

However, there are significant challenges that must be addressed before the application of such deep learning-based pose estimation algorithms in space missions. First, the SPN, trained and tested on SPEED, has shown to perform relatively poorly when the spacecraft appears too large or too small in the image\cite{Sharma2019}. Its object detection mechanism also lacked robustness when the spacecraft is occluded due to eclipse. Most importantly, neural networks are known to lack robustness to data distributions different from the one used during training, and it must be verified that these algorithms can meet the accuracy requirements on spaceborne imagery even when trained solely on synthetically generated images. This is especially challenging since spaceborne imagery can contain texture and surface illumination properties and other unmodeled camera artifacts that cannot be perfectly replicated in synthetic imagery. Since spaceborne images are expensive to acquire, the CNN must be able to address this issue with minimal or no access to the properties of spaceborne imagery.

This work makes two contributions to address the aforementioned challenges. The primary contribution of this work is a novel method to enable an efficient learning-based pose determination. Similar to SPN, the problem of pose estimation is decoupled into object detection and pose estimation networks. However, the pose estimation is performed by regressing the 2D locations of the spacecraft's surface keypoints then solving the Perspective-n-Point (PnP) problem. The extracted keypoints have known correspondences to those in the 3D model, since the CNN is trained to predict them in a pre-defined order. This design choice allows for bypassing the computationally expensive feature matching through algorithms such as RANSAC\cite{fischler_bolles_1981} and directly use publicly available PnP solvers only once per image\cite{Lepetit2008}. The proposed architecture has scored $4^\textrm{th}$ place in the recent SLAB/ESA Pose Estimation Challenge and is shown to be fast and robust to a variety of illumination conditions and inter-spacecraft separation ranging from 3 to over 30 meters. 

The secondary contribution of this work is the study of a novel training procedure that improves the robustness of the CNN to spaceborne imagery when trained solely on synthetic images. Specifically, inspired by the recent work of Geirhos et al., the technique of texture randomization is introduced as a part of the training procedure of the CNN\cite{geirhos2018imagenettrained}. Geirhos et al.~suggest that CNN tends to focus on the local texture of the target object, thus randomizing the object texture using the Neural Style Transfer (NST) technique forces the CNN to instead learn the global shape of the object\cite{jackson2018style}. Following their work, a new dataset is generated by applying NST to a custom synthetic dataset that has same pose distribution as SPEED dataset. It is shown that the network exposed to new texture-randomized dataset during training performs better on spaceborne images without having been trained on them. 

In the following section, the proposed CNN architecture is explained in detail. The section after that elaborates on the texture randomization procedure and the associated datasets used for training and validation. The section afterward introduces the experiments conducted to evaluate the performance of the proposed CNN and the effect of texture randomization. Finally, the conclusion and the directions for future work are presented.

\begin{figure}[t]
	\centering
	\includegraphics[width=0.55\textwidth]{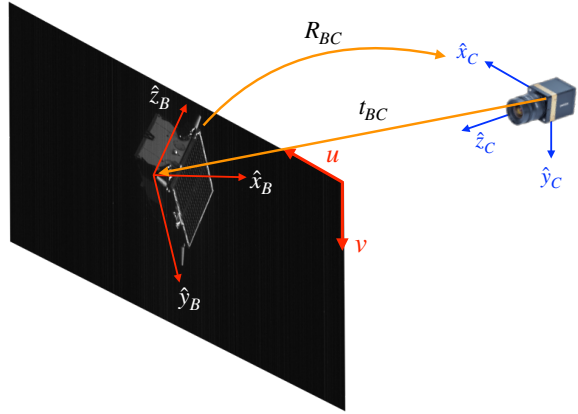}
	\caption{Definition of the body reference frame ($\mathcal{B}$), camera reference frame ($\mathcal{C}$), relative position ($\bm{t}_\mathcal{BC}$), and relative attitude ($\bm{R}_\mathcal{BC}$).}
	\label{fig:ProblemStatement}
\end{figure}

\begin{figure}[h!]
	\centering
	\includegraphics[width=0.92\textwidth]{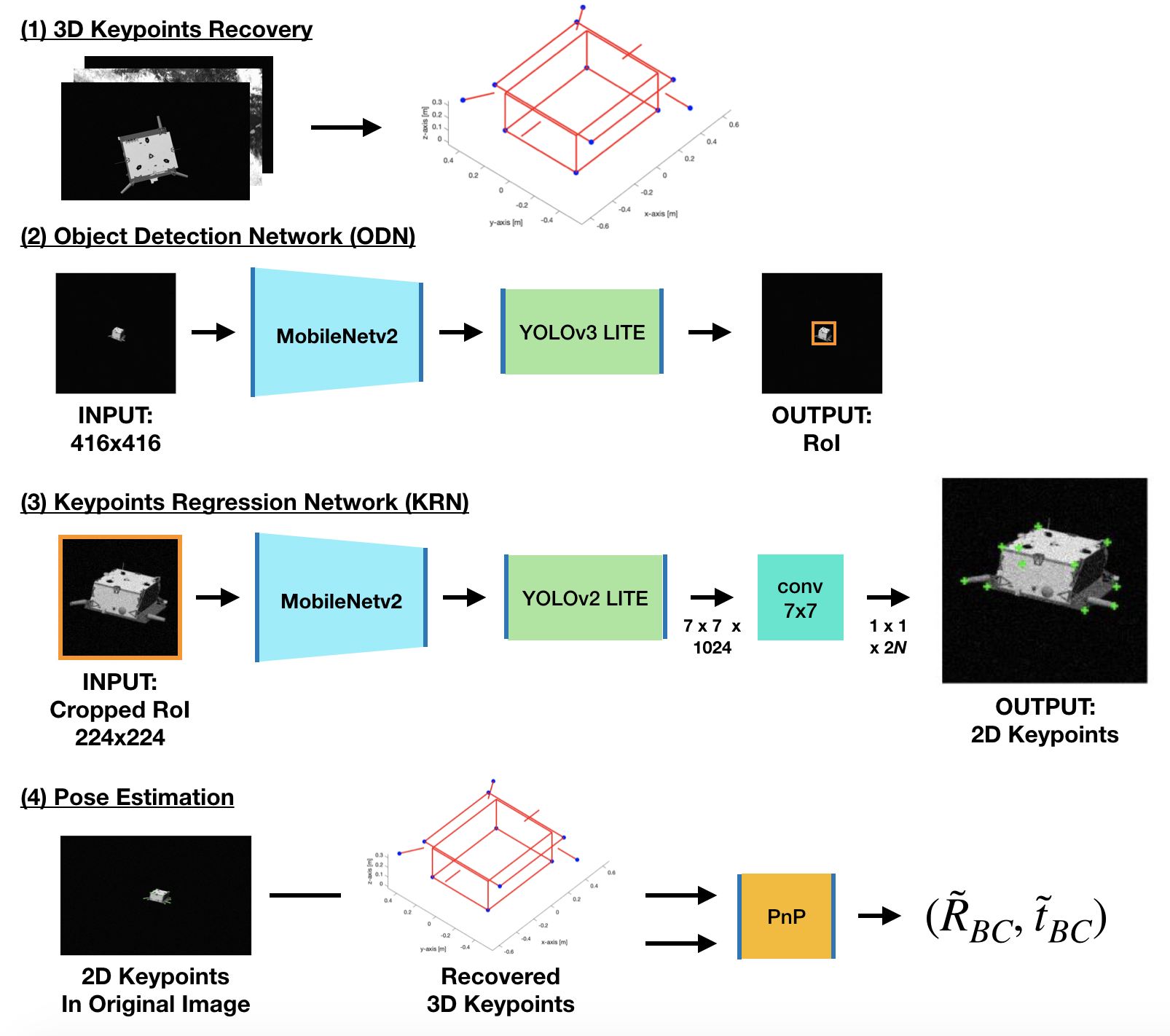}
	\caption{Overall architecture of the proposed CNN.}
	\label{fig:OverallPipelin}
\end{figure}

\section{Single Image Pose Estimation}
The general problem statement is to determine the relative attitude and position of the camera frame, $\mathcal{C}$, with respect to the target's body frame, $\mathcal{B}$. The relative position is represented by a position vector, $\bm{t}_\mathcal{BC}$, from the origin of $\mathcal{C}$ to the origin of $\mathcal{B}$. Similarly, the relative attitude is represented by a rotation matrix, $\bm{R}_\mathcal{BC}$, which aligns the reference frame $\mathcal{B}$ with $\mathcal{C}$. Figure \ref{fig:ProblemStatement} graphically illustrates these reference frames and variables.

\begin{figure}[t]
	\centering
	\includegraphics[width=0.6\textwidth]{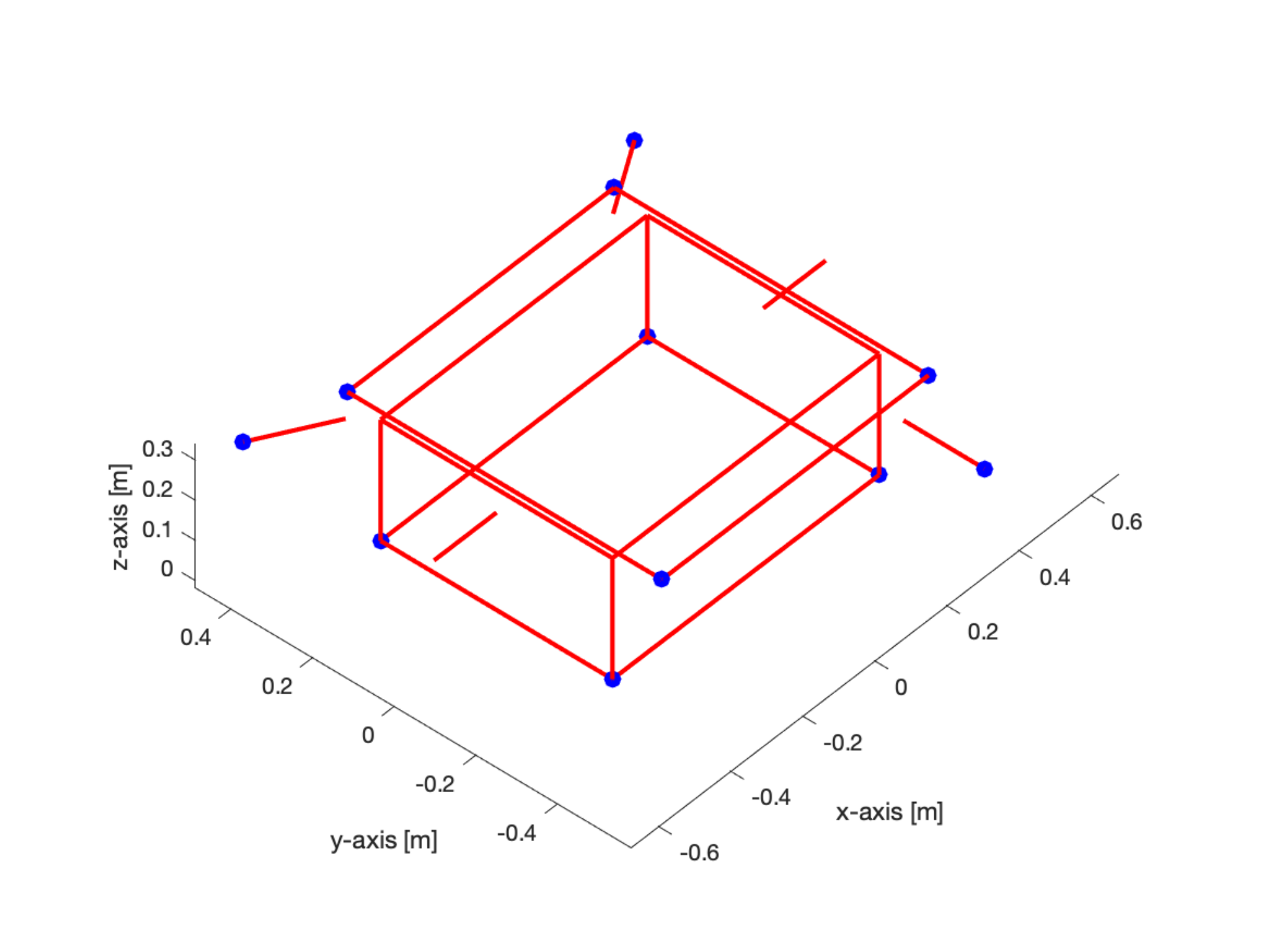}
	\caption{11 keypoints used in the proposed architecture visualized on a wireframe model of Tango spacecraft \cite{Sharma2019} (blue dots)}
	\label{fig:Tango Keypoints}
\end{figure}

The overall pipeline of the single image pose estimation architecture developed in this work is visualized in Figure \ref{fig:OverallPipelin} in four steps.
\begin{enumerate}
    \item First, the 3D model coordinates of 11 keypoints are selected from the available wireframe model of the Tango spacecraft. If the model is not available, the 3D model coordinates of the selected keypoints are recovered from a set of training images and associated pose labels. Figure \ref{fig:Tango Keypoints} visualizes the keypoints selected for this architecture, which geometrically correspond to four corners of the bottom plate, four corners of the top plate (i.e.~solar panel), and three tips of the antennae.
    \item Second, the Object Detection Network (ODN) detects a 2D bounding box around the spacecraft from the image resized to 416 $\times$ 416. The 2D bounding box labels are obtained by projecting the 3D keypoints onto an image plane using the provided ground-truth poses, then taking maximum and minimum coordinates in $x$ and $y$ directions.
    \item Third, the detected 2D bounding box is used to crop the Region-of-Interest (RoI) from the original image, which is resized to 224 $\times$ 224 and fed into the Keypoints Regression Network (KRN). The KRN returns $1 \times 2N$ vector encoding the 2D locations of $N$ keypoints. 
    \item Lastly, the extracted 2D keypoints are measured in the context of the original image. Then, they can be used in solving the PnP problem using an off-the-shelf PnP solver along with the known or recovered 3D model coordinates to compute the full 6D pose.
\end{enumerate}

\subsection{3D Keypoints Recovery}
While this work exploits the available wireframe model of the Tango spacecraft, the method of 3D keypoints recovery is introduced for completeness. In order to recover the 3D model coordinates ($\bm{p}_\textrm{3D}$) of the aforementioned 11 keypoints, a set of 12 training images is selected in which the Tango spacecraft is well-illuminated and has varying poses. Then, a set of visible keypoints ($\bm{p}_\textrm{2D}$) is manually picked. In order to recover $\bm{p}_\textrm{3D}$, the following optimization problem is solved,
\begin{equation} \label{eqn:3D keypoint reconstruction}
\textrm{minimize} ~ \sum_j || s_j \bm{p}_{\textrm{2D}, k}^h - \bm{K}[ \bm{R}_j | \bm{t}_j ] \bm{p}_{\textrm{3D},k}^h ||_2
\end{equation}
where, for each $k$-th point, the sum of the reprojection error is minimized over a set of images in which the $k$-th point is visible. In Eq.~(\ref{eqn:3D keypoint reconstruction}), a superscript $h$ indicates the point is expressed in homogenous coordinates, $\bm{K}$ is a known camera intrinsic matrix, and ($\bm{R}_j, \bm{t}_j$) is a known pose associated with the $j$-th image. The optimization variables in Eq.~(\ref{eqn:3D keypoint reconstruction}) are the 3D model coordinates, $\bm{p}_{\textrm{3D},k}$, and scaling factors, $s_j$, associated with the projection onto the image plane in each input image. Since Eq.~(\ref{eqn:3D keypoint reconstruction}) is a convex objective function of its optimization variables, the solutions, $(\bm{p}_{\textrm{3D},k}^h, s_j) ~\forall~ j = 1,\dots,11$, are obtained using the CVX solver \cite{cvx, gb08}.

\begin{figure}[h!]
	\centering
	\includegraphics[width=0.6\textwidth]{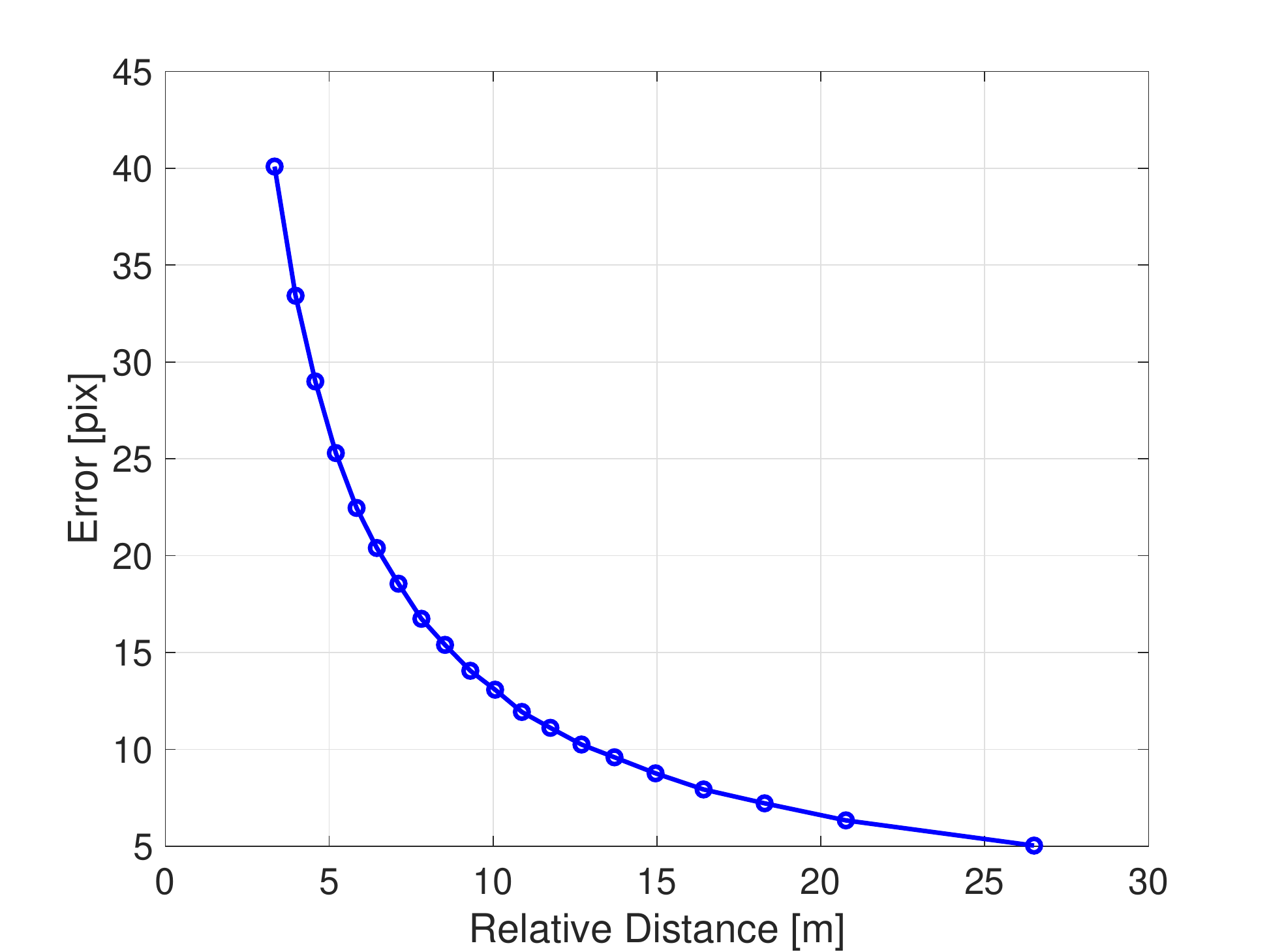}
	\caption{Average of reprojection error of recovered 3D keypoints plotted against mean relative distance, $||\bm{t}_\mathcal{BC}||_2$, for the SPEED synthetic training images.}
	\label{fig:Keypoint Projection}
\end{figure}

Overall, the reconstructed 3D keypoint coordinates have an average distance error of 5.7 mm compared to those in the wireframe model, with the maximum error around 9.0 mm. Figure \ref{fig:Keypoint Projection} plots the reprojection error of the recovered keypoints against the ground-truth keypoints from the wireframe model. While the maximum average error is around 40 pixels, the CNN trained with labels from recovered keypoints implicitly learns the offsets from the ground-truth coordinates.

\begin{figure}[h!]
	\centering
	\includegraphics[width=0.5\textwidth]{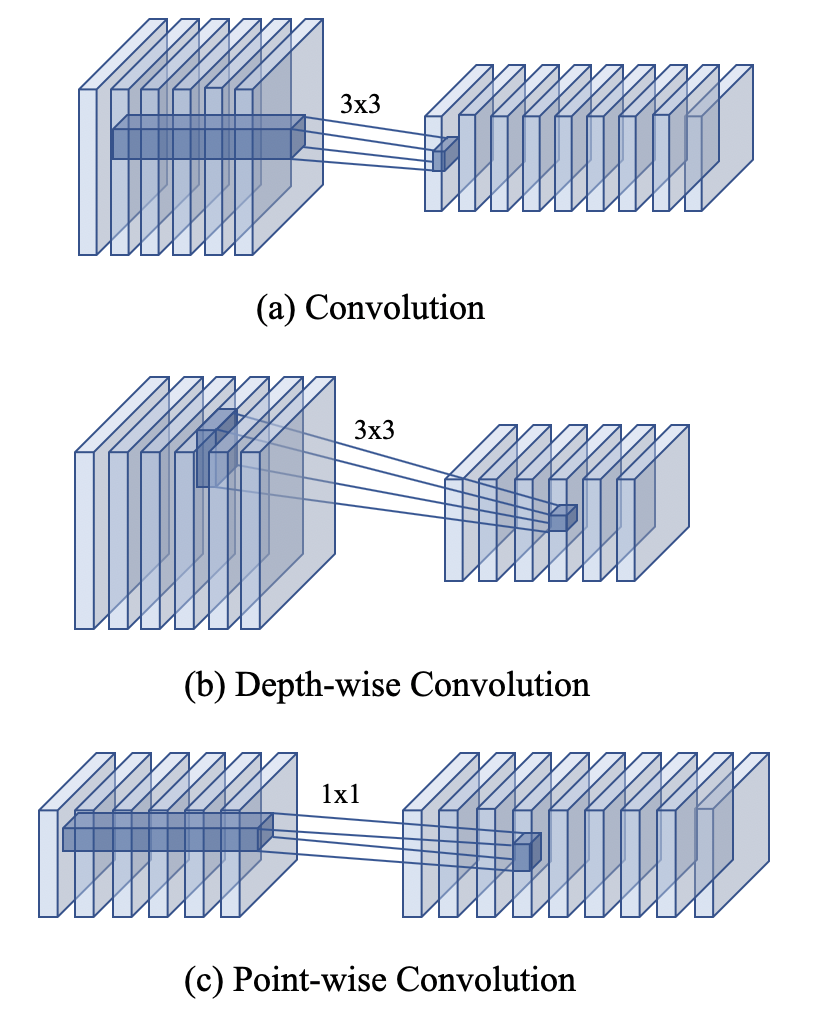}
	\caption{Different convolution operations. In this work, conventional convolution operation (a) is replaced by depth-wise convolution (b) followed by the point-wise convolution (c).}
	\label{fig:Convolutions}
\end{figure}
\subsection{Object Detection Network (ODN)}
The ODN pipeline closely follows the structure of the state-of-the-art detection network, YOLOv3 \cite{redmon_farhadi_2018}. It takes an input of $416 \times 416$ and performs detection at three different stages. The original backbone of Darknet-53 and extra convolutional layers are replaced with MobileNetv2\cite{Sandler_MobileNet_v2} and depth-wise separable convolution operations\cite{howard2017mobilenets}, respectively, to drastically reduce the number of parameters in the network. Specifically, depth-wise separable convolution breaks up the conventional convolution into depth-wise and point-wise convolution operations. As the names suggest, depth-wise convolution applies a single kernel per each channel, compared to conventional convolution that applies kernels over all channels. On the other hand, point-wise convolution applies a 1$\times$1 kernel over all channels and, unlike depth-wise convolution, can be used to output arbitrary number of channels. Figure \ref{fig:Convolutions} qualitatively describes different convolution operations. As in MobileNet\cite{howard2017mobilenets}, both depth-wise and point-wise convolutions are followed by batch normalization and Rectified Linear Unit (ReLU) activation function layers.

In general, for a convolution unit with kernel size $K \times K$, $C_\textrm{in}$ input channels and $C_\textrm{out}$ output channels, the number of tunable parameters is given as 
\begin{equation}
    K \times K \times C_\textrm{in} \times C_\textrm{out},
\end{equation}
whereas depth-wise separable convolution has
\begin{equation}
    K \times K \times C_\textrm{in} \times 1 + 1 \times 1 \times C_\textrm{in} \times C_\textrm{out}
\end{equation}
parameters. The factor of reduction in number of parameters is then given as
\begin{equation}
    \frac{1}{C_\textrm{out} } + \frac{1}{K \times K}.
\end{equation}
Given that the most common kernel size in state-of-the-art CNN architectures is $3 \times 3$, simply replacing each convolution with depth-wise separable convolution reduces computation by a factor of 8 or 9.

Since SPEED guarantees the presence of a single, known spacecraft in every image, no classification is performed in the ODN. The direct output of the ODN is an $N \times N \times 5$ tensor, where $N$ = 13, 26, 52 in respective prediction stages. The output tensor essentially divides the input image into $N \times N$ grids, each grid predicting $(t_0, t_x, t_y, t_w, t_h)$. These predictions are related to the objectness score, $p(c)$, the location of the bounding box center, $(x, y)$, and the size of the bounding box, $(w, h)$, via the following equations,
\begin{align}
    p(c) &= \sigma(t_0) \nonumber \\
    x    &= \sigma(t_x) + g_x \nonumber \\
    y    &= \sigma(t_y) + g_y \\
    w    &= p_w e^{t_w} \nonumber \\
    h    &= p_h e^{t_h} \nonumber 
\end{align}
where $\sigma(x)$ is a sigmoid function, $(g_x, g_y)$ is the location of each grid, and $(p_w, p_h)$ is the size of each anchor box. Similar to YOLOv3, a total of nine anchor boxes are pre-defined using k-means clustering, and three are used for prediction in each stage. The ground-truth objectness score is assigned 1 to a grid containing the object and the best-matching anchor. Since there is only one object, the prediction is made from the grid with the highest objectness score during the inference without non-max suppression. The readers are encouraged to refer to a series of publications on YOLO architecture for more details on the implementation\cite{Redmon2016, redmon_farhadi_2017, redmon_farhadi_2018}. 

Compared to the original YOLOv3, the loss function is modified to better maximize the Intersection-over-Union (IoU) metric defined as
\begin{equation}\label{eqn:iou}
    \textrm{IoU} = \frac{I}{U} = \frac{\textrm{Area of Intersection}}{\textrm{Area of Union}}.
\end{equation}
Specifically, the Mean-Squared Error (MSE) loss of the bounding box parameters is replaced by the Generalized Intersection-over-Union (GIoU) loss defined as \cite{Rezatofighi2018_GIOU}
\begin{equation}
    \mathcal{L}_\textrm{GIoU} = 1 - \textrm{GIoU}, ~\textrm{where}~ \textrm{GIoU} = \frac{I}{U} - \frac{A^\textrm{C} - U}{A^\textrm{C}},
\end{equation}
where $A^\textrm{C}$ is an area of the smallest box enclosing both predicted and ground-truth bounding boxes. The formulation of GIoU loss ensures that the gradient is bigger as the separation between bounding boxes becomes larger even when there is no overlap (i.e.~IoU =
0). The overall loss, excluding the classification loss, is then
\begin{equation}
    \lambda_\textrm{GIoU}\mathcal{L}_\textrm{GIoU} + \lambda_\textrm{conf}\mathcal{L}_\textrm{conf}
\end{equation}
where $\lambda_\textrm{GIoU}$ and $\lambda_\textrm{conf}$ are weighting factors, and $\mathcal{L}_\textrm{conf}$ is a sum of the Binary Cross-Entropy (BCE) loss between predicted and ground-truth objectness score.

\subsection{Keypoints Regression Network (KRN)}
The input to the KRN is a RoI cropped from the original image using the 2D bounding box detected from the ODN. The motivation behind the cropping is the fact that the SPEED images are large (1920 $\times$ 1200 pixels) compared to the input sizes typically demanded by CNN architectures (e.g.~224 $\times$ 224 for VGG-based networks). Regular resizing from 1920 $\times$ 1200 to 224 $\times$ 224 will blur much of the detailed features that can help make accurate prediction of the keypoint locations, especially if the target appears small due to large inter-spacecraft separation. Therefore, cropping the RoI prior to KRN helps the network make better predictions based on much finer features. In general, this approach works regardless of the image size and makes the architecture robust to different feature resolutions.

The structure of the KRN closely follows the architecture of YOLOv2 but exploits the MobileNet architecture and depth-wise separable convolution operations similar to ODN. It receives the input, which is cropped around the RoI and resized to 224 $\times$ 224, and outputs a 7 $\times$ 7 $\times$ 1024 tensor. The output tensor is reduced to a 1 $\times$ 2$N$ vector via a convolution with 7 $\times$ 7 kernel to regress the 2D locations of the $N$ keypoints, where $N$ = 11 as defined earlier. It is empirically found that dimension reduction using a convolution performs better than using global average pooling. These keypoints are then used to compute the 6D pose estimate using the EPnP algorithm \cite{Lepetit2008} with the selected 3D keypoint coordinates. The loss function of the KRN is simply a sum of MSE between the predicted ($\tilde{\bm{k}}$) and ground-truth keypoint ($\bm{k}$) locations, i.e.
\begin{equation}
    \mathcal{L}_\textrm{KRN} = \sum_{j=1}^{11} ||\tilde{k}_x^{(j)} - k_x^{(j)}||_2 + ||\tilde{k}_y^{(j)} - k_y^{(j)}||_2.
\end{equation}

\section{Texture Randomization}

The advantage of many benchmark datasets for various deep learning-based computer vision tasks, such as ImageNet\cite{Krizhevsky2012}, MS COCO\cite{Lin2014COCO}, or LINEMOD\cite{Hinterstoisser2013_LINEMOD}, is that they comprise images from the real world to which the CNNs are expected to be applied. However, due to the difficulty of acquiring the same amount of spacecraft images with accurately annotated labels, the training dataset for spaceborne CNN inevitably depends heavily on synthetic renderers and laboratory testbeds. Unfortunately, it is extremely difficult to exactly replicate the target spacecraft's surface properties and lighting conditions encountered throughout a space mission. Since any CNN for space application is likely to be trained mainly on a set of synthetic images, the gap between the properties of synthetic and spaceborne images must be addressed to ensure the CNN's functionality in space missions. While it is possible to include some spaceborne images during training to improve the CNN's generalizability, this work specifically considers only the availability of synthetic images during training.

Arguably, one of the most distinct differences between the synthetic and spaceborne images is surface texture. Recently, Geirhos et al.~have empirically shown that many state-of-the-art CNN architectures, such as ResNet\cite{He2015_ResNet} and VGG\cite{Simonyan2014}, exhibit strong bias toward object's texture when trained on ImageNet\cite{geirhos2018imagenettrained}. This finding is very counter-intuitive for humans who would, for example, classify an object or an animal based on its global or a set of local shapes, not based on its texture. To demonstrate such behavior, the authors have created Stylized-ImageNet (SIN) dataset by applying the AdaIN Neural Style Transfer (NST) pipeline\cite{huang2017adain} to each image in ImageNet with random styles from Kaggle's \texttt{Painter by Numbers} dataset\footnote{https://www.kaggle.com/c/painter-by-numbers/}. The result shows that when the same CNN architectures are trained instead on SIN dataset, the networks not only exceed the performance of those trained on ImageNet, they also show human-level robustness to previously unseen image distortions, such as noise, contrast change, and high- or low-pass filtering.

In this work, a similar approach is adopted to emphasize the significance of the spacecraft texture on the CNN's performance. First, a synthetic dataset PRISMA12K is created using the same rendering software used in SPEED. PRISMA12K consists of 12,000 synthetic images of the Tango spacecraft with the same pose distribution as SPEED. However, PRISMA12K uses the camera model of the vision-based sensor used on the Mango spacecraft during the PRISMA mission (Table \ref{tab:PRISMAcamera}).

\begin{table}[h!]
	\caption{PRISMA camera parameters and values.}
	\label{tab:PRISMAcamera}
	\centering
	\begin{tabular}{@{}ccc@{}} \toprule
	    Parameter & Description & Value \\ \midrule
	    $N_\textrm{u}$ & Number of horizontal pixels & 752 \\
	    $N_\textrm{v}$ & Number of vertical pixels & 580 \\
	    $f_\textrm{x}$ & Horizontal focal length & 0.0200 m \\
	    $f_\textrm{v}$ & Vertical focal length & 0.0193 m \\
	    $du$ & Horizontal pixel length & $8.6 \times 10^{-6}$ m \\
	    $dv$ & Vertical pixel length & $8.3 \times 10^{-6}$ m \\
		\bottomrule
	\end{tabular}
\end{table}

\begin{figure}[t]
    \centering
    \includegraphics[width=\textwidth]{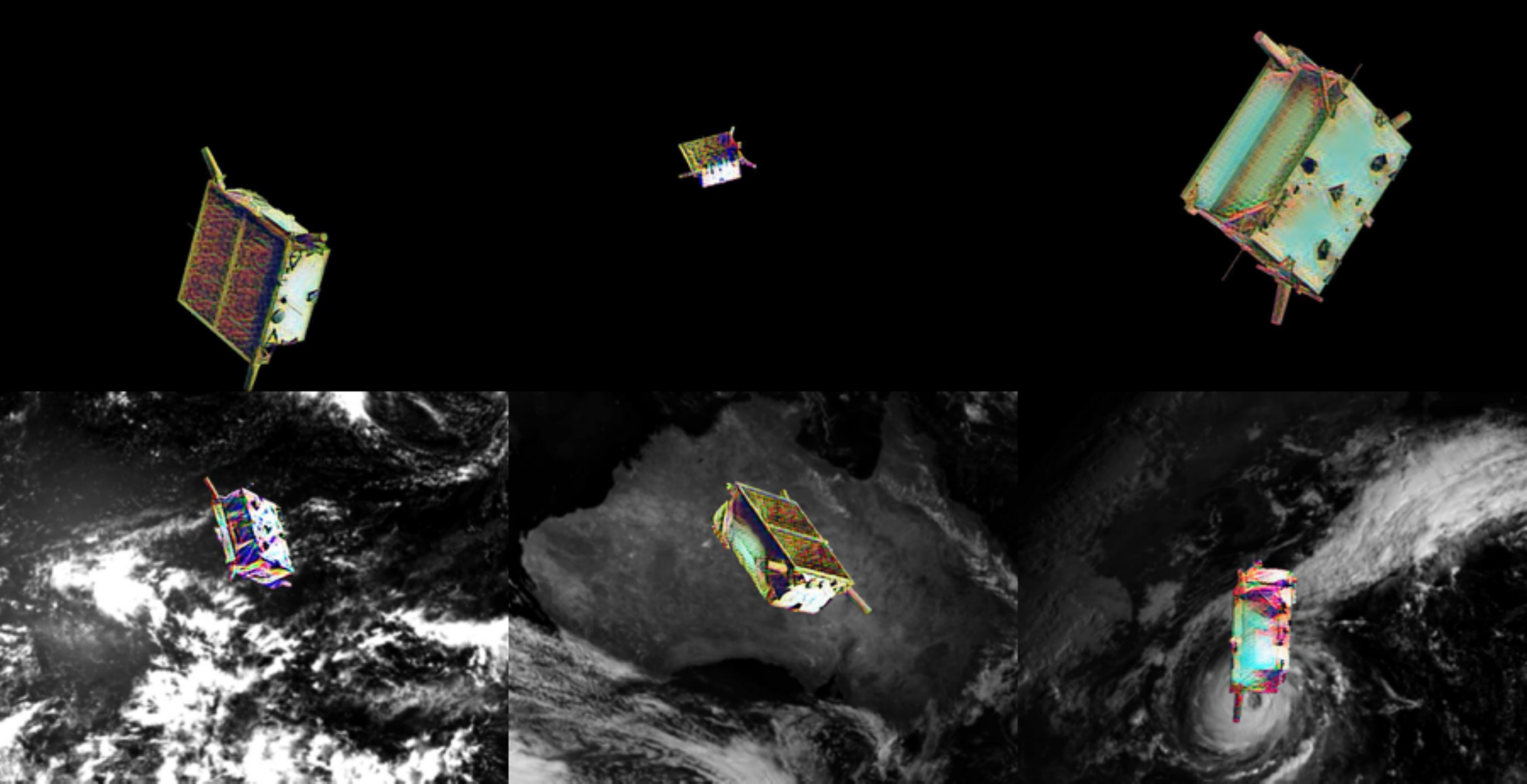}
    \caption{Examples of 6 images from PRISMA12K-TR}
    \label{fig:speed_stylized}
\end{figure}

A second dataset, PRISMA12K-TR, is created by applying a NST pipeline to each image of the PRISMA12K offline to randomize the spacecraft texture. In this work, the pre-trained NST pipeline proposed by Jackson et al.~is used \cite{jackson2018style}. Instead of explicitly supplying the style image at each inference, this NST pipeline allows for randomly sampling a vector of style embedding $z \in \mathcal{R}^{100}$. Specifically, the style embedding is sampled as
\begin{equation}
    z = \alpha \mathcal{N}(\mu, \Sigma) + (1 - \alpha)P(c)
\end{equation}
where $P(c)$ is the style embedding of the content image, ($\mu, \Sigma$) are the mean vector and covariance matrix of the style image embeddings pre-trained on ImageNet, and $\alpha$ is the strength of the random normal sampling. In this work, $\alpha = 0.25$ is used to create PRISMA12K-TR. In order to avoid the NST's blurring effect on the spacecraft's edges, the style-randomized spacecraft is first isolated from the background using a bitmask then combined with the original background. Figure \ref{fig:speed_stylized} shows a montage of six such images.

The third dataset is PRISMA25, which consists of 25 spaceborne images captured during the rendezvous phase of the PRISMA mission\cite{PRISMA_chapter}. The PRISMA25 is used to evaluate the performance of the CNN on a previously unseen spaceborne dataset when trained solely on a mixture of PRISMA12K and PRISMA12K-TR. 


\section{Experiments}

In this section, the procedures and results of two experiments are elaborated. Throughout both experiments, two variants of the keypoint regression network are trained and evaluated. The first variant, noted as KRN-SK, is identical to the KRN introduced in Secion 2 and regresses the 2D coordinates of 11 surface keypoints. The second variant, noted as KRN-BB, instead regresses the 2D coordinates of the centroid and eight corners of the 3D bounding box around the object. 

The first experiment evaluates the performance of the proposed single image pose estimation architecture, namely ODN and KRN-SK. In order to provide an in-depth analysis, both networks are first trained on 80\% of the 12,000 synthetic training images and evaluated on the validation set which comprises the rest of the 20\%. The performance of the combined architecture on synthetic and real test sets is also reported. The second experiment instead trains KRN-SK and KRN-BB using mixtures of PRISMA12K and PRISMA12K-TR using the ground-truth 2D bounding boxes. Both versions of KRN are evaluated on PRISMA25 for comparison and in order to gauge the effect of texture randomization in closing the domain gap between synthetic and spaceborne images. The keypoint labels are generated using the ground-truth wireframe model of the Tango spacecraft unless stated otherwise.

\subsection{Evaluation Metrics}
Throughout this section, four performance metrics are used to evaluate the proposed architecture. For ODN, the mean and median IoU scores are reported as in Eq.~(\ref{eqn:iou}) to measure the degree of overlap between the predicted and ground-truth 2D bounding boxes. For the combined architecture, mean and median translation and rotation errors are reported as\cite{Sharma2019}
\begin{equation}
    \bm{E_\textrm{T}} = | \tilde{\bm{t}}_\mathcal{BC} - \bm{t}_\mathcal{BC}|,
\end{equation}
\begin{equation}
    E_\textrm{R} = 2 \arccos |\bm{q}_\mathcal{BC} \cdot \tilde{\bm{q}}_\mathcal{BC} |
\end{equation}
where $(\tilde{\bm{q}}_\mathcal{BC}, \tilde{\bm{t}}_\mathcal{BC})$ are predicted unit quaternion and translation vector aligning the target body frame ($\mathcal{B}$) and camera frame ($\mathcal{C}$), and $(\bm{q}_\mathcal{BC}, \bm{t}_\mathcal{BC})$ are ground-truth unit quaternion and translation vector. Lastly, the pose score used in SLAB/ESA Pose Estimation Challenge (henceforth noted SLAB/ESA score) is reported as
\begin{equation}
    \textrm{SLAB/ESA score} = \frac{1}{N}\sum_{i=1}^N \frac{||\tilde{\bm{t}}_\mathcal{BC}^{(i)} - \bm{t}_\mathcal{BC}^{(i)}||_2}{||\bm{t}_\mathcal{BC}^{(i)}||_2} + E_\textrm{R}^{(i)}.
\end{equation}

\subsection{Experiment 1: Single Image Pose Estimation}
For single image pose estimation, both ODN and KRN are trained using the RMSprop optimizer \cite{Tieleman2012} with batch size of 48 and momentum and weight decay set to 0.9 and $5 \times 10^{-5}$, respectively, unless stated otherwise. For both networks, the learning rate is initially set to 0.001 and decays exponentially by a factor of 0.98 after every epoch. The networks are implemented with PyTorch v1.1.0 and trained on an NVIDIA GeForce RTX 2080 Ti 12GB GPU for 100 epochs for ODN and 300 for KRN. No real images are used in training to gauge the architecture's ability to generalize to the datasets from different domains.

\begin{table}[h!]
	\caption{Parameters and distributions of data augmentation techniques. Brightness, contrast, and Gaussian noise are implemented with 50\% chance during training.}
	\label{tab:data augmentation}
	\centering
	\begin{tabular}{@{}ccccc@{}} \toprule
		Brightness ($\beta$) & Contrast ($\alpha$) & Gaussian Noise & \makecell{ RoI Enlargement \\ Factor [\%]} & \makecell{RoI Shifting \\ Factor [\%]} \\ \midrule
        $\mathcal{U}(-25, 25)$ & $\mathcal{U}(0.5, 2.0)$ & $\mathcal{N}(0, 25)$ & $\mathcal{U}(0, 50)$ & $\mathcal{U}(-10, 10)$\\
		\bottomrule
	\end{tabular}
\end{table}

In training both networks, a number of data augmentation techniques is implemented. For both ODN and KRN, the brightness and contrast of the images are randomly changed according to
\begin{equation}
    p^\prime(i,j) = \alpha p(i,j) + \beta
\end{equation}
where $p(i,j) \in [0, 255]$ is the value of a pixel at $i^\textrm{th}$ column and $j^\textrm{th}$ row of the image. The images are randomly flipped and rotated at 90$^\circ$ intervals, and a random Gaussian noise is also implemented. For KRN, the ground-truth RoI is first corrected to a square-sized region with a size of $\max(w, h)$, where $(w, h)$ are the width and height of the original RoI. This correction is implemented to ensure the aspect ratio remains the same when resizing the cropped region into 224 $\times$ 224. Then, the new square RoI is enlarged by a random factor up to 50\% of the original size. Afterwards, the enlarged RoI is shifted in horizontal and vertical directions by a random factor up to 10\% of the enlarged RoI dimension. This technique has an effect of making the network robust to object translation and misaligned RoI detection. During testing, the detected RoI is similarly converted into a square-sized region and enlarged by a fixed factor of 20\% to ensure the cropped region contains the entirety of the spacecraft. The distributions of each augmentation parameter are summarized in Table \ref{tab:data augmentation}.

\begin{table}[t]
	\caption{Performance of the proposed architecture on synthetic validation set.}
	\label{tab:SPEED_valid}
	\centering
	\begin{tabular}{@{}cc@{}} \toprule
		Metrics & SPEED Synthetic Validation Set \\ \midrule
		Mean IoU & 0.919 \\
		Median IoU & 0.936 \\
		Mean $\bm{E_\textrm{T}}$ [m] & [ 0.010, 0.011, 0.210] \\
		Median $\bm{E_\textrm{T}}$ [m] & [0.007, 0.007, 0.124] \\
		Mean $E_\textrm{R}$ [deg]  & 3.097 \\
		Median $E_\textrm{R}$ [deg] & 2.568 \\
		SLAB/ESA Score & 0.073 \\
		\bottomrule
	\end{tabular}
\end{table}
Table \ref{tab:SPEED_valid} reports the proposed CNN's performance on the SPEED synthetic validation dataset. Overall, the ODN excels in detecting the spacecraft from the images with mean IoU of 0.919. The worst IoU score in the validation set is reported as 0.391, indicating that even the worst prediction of the 2D bounding box still has some overlap with the target, mitigating the effect of misaligned RoI on the keypoints regression. The pose solutions of the combined architecture also show improved performance on synthetic validation set compared to that of SPN\cite{Sharma2019}. Specifically, the mean $E_\textrm{T}$ is under 25 cm, while the mean $E_\textrm{R}$ is around 3.1 degrees. 

\begin{figure}[p!]
    \centering
    \includegraphics[width=\textwidth]{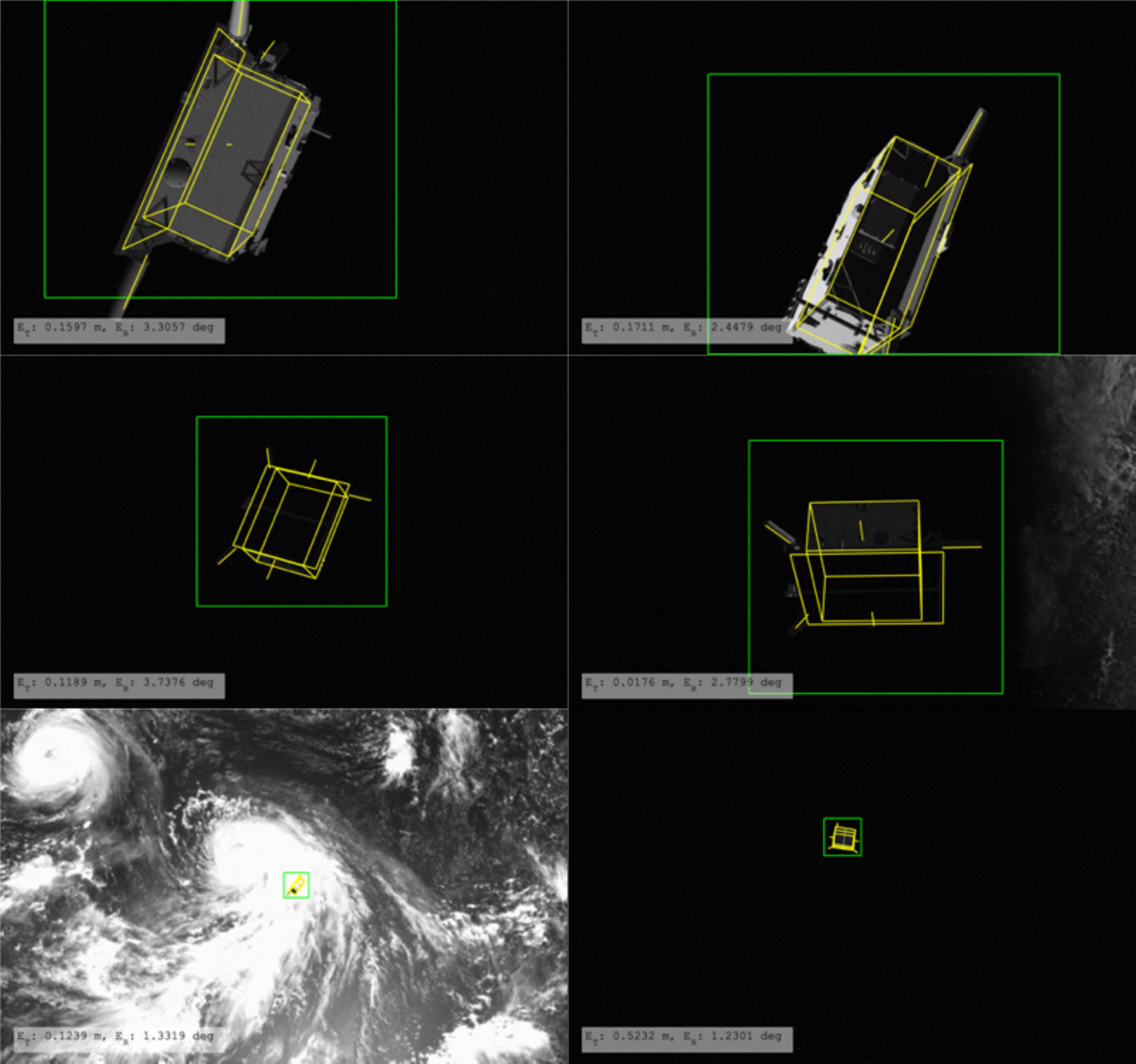}
    \caption{Examples of the predicted 2D bounding boxes and pose solutions of the proposed architecture on the SPEED validation set. The 2D bounding boxes shown are enlarged by 20\% from the predicted boxes. In general, the proposed CNN excels in the cases of proximity (top), severe shadowing (middle), and large separation (bottom).}
    \label{fig:speed_goodresults}
\end{figure}

\begin{figure}[t]
    \centering
    \includegraphics[width=\textwidth]{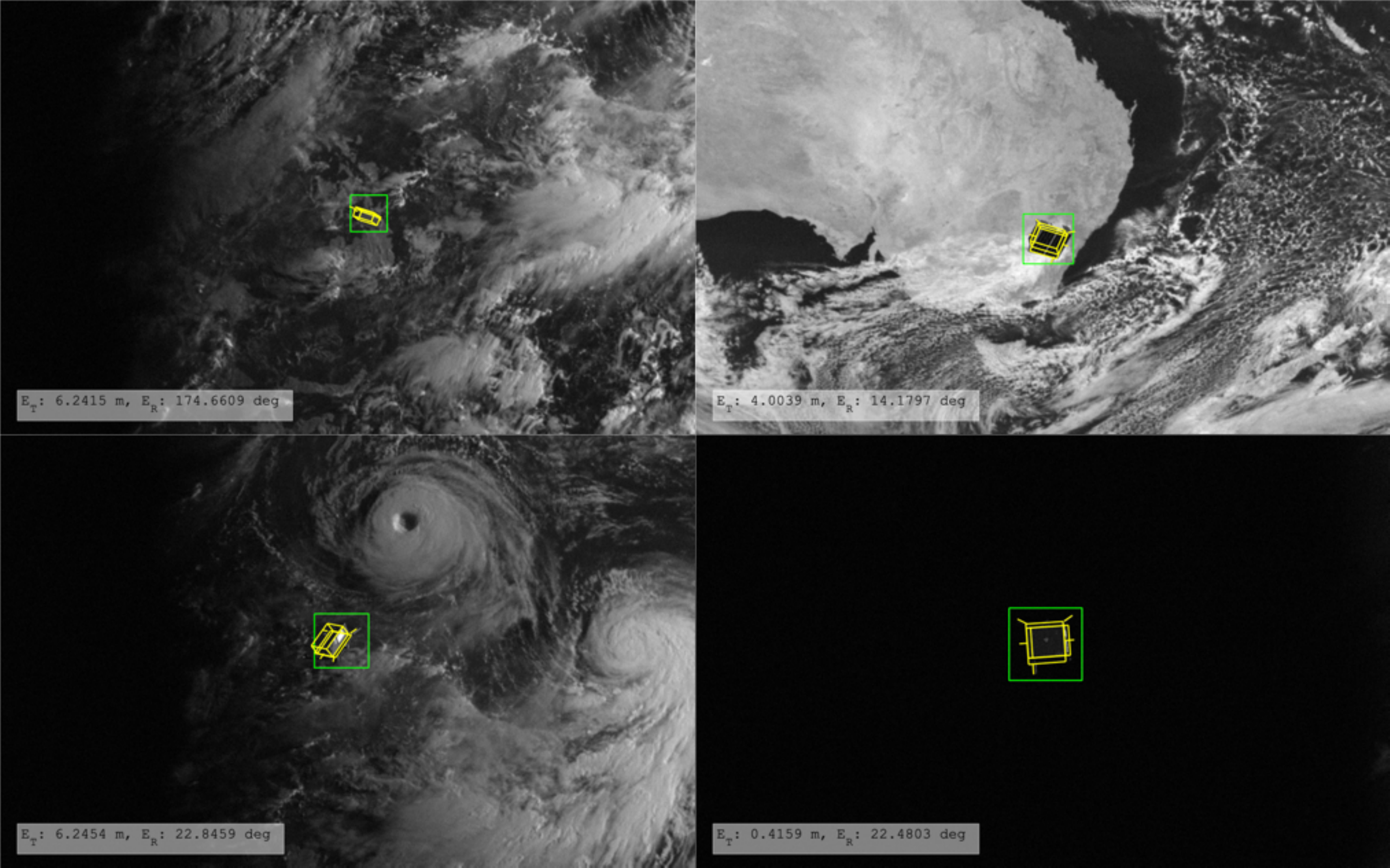}
    \caption{Four worst pose solutions on the SPEED validation set. The 2D bounding boxes shown are enlarged by 20\% from the predicted boxes. Even with cropping around the RoI, in some cases the spacecraft features blend into the Earth background (top) or Gaussian noise (bottom-right).}
    \label{fig:speed_badresults}
\end{figure}

Figure \ref{fig:speed_goodresults} visualizes six cases of successful pose predictions. Overall, the figures demonstrate both ODN and KRN are able to make accurate predictions despite clipping due to proximity, severe shadowing, or large inter-spacecraft separation above 30 meters regardless of the presence of Earth in the background. However, as shown in the cases of the four worst predictions visualized in Figure \ref{fig:speed_badresults}, the CNN is not always immune to large separation or extreme shadowing. Figure \ref{fig:speed_badresults} demonstrates that the combined networks, despite the ``zooming-in'' effect from cropping around the RoI, can still fail in accurate keypoints regression when the inter-spacecraft separation is too large because the keypoint features of the spacecraft become indistinguishable from the Earth in the background. In other case, the boundary of the spacecraft's shape blurs and blends into the Gaussian noise in the background due to the shadowing and large separation. On the other hand, the ODN is able to predict the bounding boxes with high accuracy even in the four worst prediction cases.

\begin{figure}[h!]
    \centering
    \includegraphics[width=0.49\textwidth]{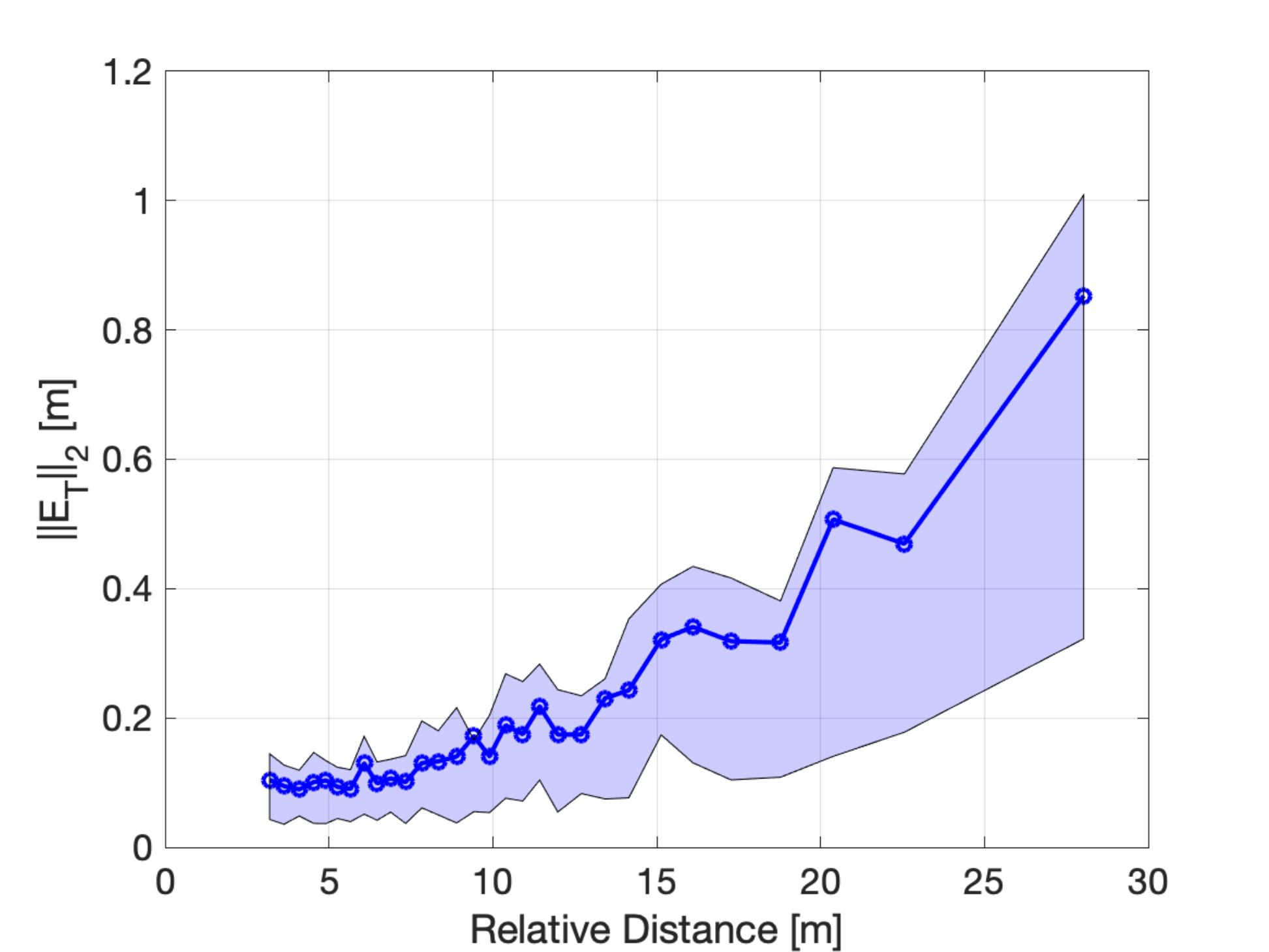}
    \includegraphics[width=0.49\textwidth]{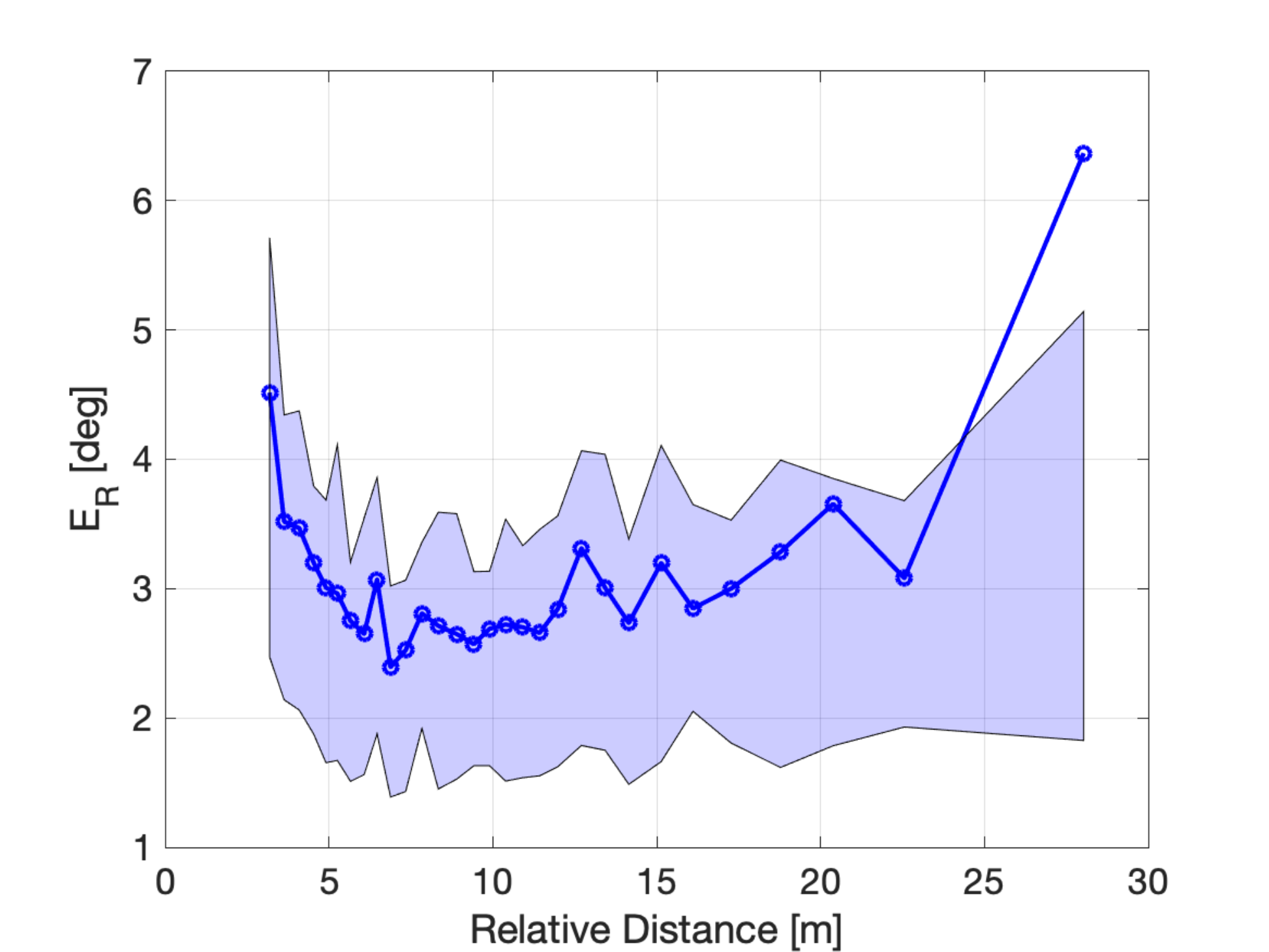}
    \caption{Mean $||E_\textrm{T}||_2$ and $E_\textrm{R}$ plotted against mean relative distance, $||\bm{t}_\mathcal{BC}||_2$, for the SPEED validation set. The shaded region shows 25 and 75 percentile values.}
    \label{fig:speedscore}
\end{figure}

Figure \ref{fig:speedscore} plots the average of the translation and rotation errors with respect to the mean ground-truth relative distance. The distribution of the errors exhibits the trend also visible in SPN\cite{Sharma2019} -- the position error grows as the spacecraft is farther away, and the rotation error is worse when spacecraft is too close or too far. Specifically, the mean rotation error for the largest inter-spacecraft distance suffers from extreme outliers, such as the case visualized in the top-left figure of Figure \ref{fig:speed_badresults}. However, in general, the mean translation error is under a meter due to the successful performance of ODN for all range of inter-spacecraft separation, and unlike SPN, the clipping due to proximity does not cause spike in translation error thanks to the random RoI shifting during training.

\begin{table}[h!]
	\caption{Size and speed of the proposed architecture. Inference speed only accounts for the forward propagation of the ODN and KRN without any post-processing steps. }
	\label{tab:ODN_KRN_specs}
	\centering
	\begin{tabular}{@{}ccccc@{}} \toprule
		Network & \makecell{Number of parameters \\\relax [Millions]} & \makecell{Size \\\relax [MB]} & \makecell{Runtime on \\ GPU [ms]} & \makecell{Runtime on \\ CPU [ms]} \\ \midrule
		ODN & 5.53 & 22.4 & 7 & 230 \\
		KRN & 5.64 & 22.8 & 7 & 32 \\
		Total & 11.17 & 45.2 & 14 & 262 \\
		\bottomrule
	\end{tabular}
\end{table}

Table \ref{tab:ODN_KRN_specs} lists the number of parameters and inference time associated with each network in the proposed architecture. Due to the MobileNet architecutre and innovative depth-wise convolution operations, the proposed architecture requires less computation despite exploiting the architectures of the state-of-the-art deep networks. For example, the YOLOv2-based KRN only has 5.64 million parameters compared to about 50 million of the YOLOv2. By itself, the KRN runs at 140 Frames Per Second (FPS) on GPU and about 30 FPS on an Intel$\textregistered$ Core$\texttrademark$ i9-9900K CPU at 3.60GHz for inference. Similar trend can be observed for the YOLOv3-based ODN; however, the inference time on CPU increases dramatically most likely due to the upsampling operations inherent to the YOLOv3 architecture. Overall, the combined architecture runs at about 70 FPS on a GPU and 4 FPS on a CPU. An architecture like MobileNet can potentially pave the way towards the implementation of deep CNN-based algorithms on-board the spacecraft with limited computing resources.

\begin{table}[h!]
	\caption{Performance of the proposed architecture on SPEED test sets. In this case, the recovered keypoints are used as labels during training.}
	\label{tab:SPEED_testscore}
	\centering
	\begin{tabular}{@{}ccc@{}} \toprule
		Metric & SPEED Synthetic Test Set & SPEED Real Test Set \\ \midrule
		SLAB/ESA Score & 0.0626 & 0.3951 \\
		Placement & 4th & 4th \\
		\bottomrule
	\end{tabular}
\end{table}

The SLAB/ESA scores on both synthetic and real \emph{test} sets are also reported in Table \ref{tab:SPEED_testscore}. In this case, both networks are trained with all 12,000 synthetic training images. It is clear that with a bigger training dataset, the score decreases compared to that reported in Table \ref{tab:SPEED_valid}. With the synthetic score of 0.0626, the proposed architecture has scored $4^\textrm{th}$ place in the SLAB/ESA Pose Estimation Challenge. However, because the training only involved the synthetic images, the score on the real test set is about six times worse than that on the synthetic test set. 

\subsection{Experiment 2: Texture Randomization}

For texture randomization, only the performances of the KRNs are tested, as the object detection has shown to improve with texture-randomized training images from the literature\cite{jackson2018style}. In this experiment, AdamW optimizer\cite{loshchilov2019decoupled} is used with momentum and weight decay set to 0.9 and 5 $\times$ $10^{-5}$, respectively. The learning rate is initially set to 0.0005 and halves after every 50 epochs. Both KRN-SK and KRN-BB are trained for 200 epochs on the same GPU hardware as introduced in Experiment 1. 

Both variants of the KRN are trained using the ground-truth RoI that are randomly enlarged and shifted similar to the first experiment. For each input image, the network chooses an image from PRISMA12K-TR over PRISMA12K with probability of $p_\textrm{TR}$. For images from PRISMA12K, the same data augmentation techniques in Table \ref{tab:data augmentation} are used, except Gaussian noise is sampled from $\mathcal{N}(0, 10)$. For images from PRISMA12K-TR, random erasing augmentation technique\cite{zhong2017random} is applied with 50\% probability in order to mimic the shadowing effect due to the eclipse. This is because the NST cancels any illumination effect that was cast on the spacecraft, as seen in Figure \ref{fig:speed_stylized}.

\begin{table}[h!]
	\caption{Performance of the KRN-BB on PRISMA25 for varying $p_\textrm{TR}$. The average and standard deviation of SLAB/ESA score over 3 tests are reported for the best and last epochs. Bold face numbers represent the best performances.}
	\label{tab:performance_a25}
	\centering
	\begin{tabular}{@{}ccccc@{}} \toprule
		{} & $p_\textrm{TR}$ = 0 & $p_\textrm{TR}$ = 0.25 & $p_\textrm{TR}$ = 0.50 & $p_\textrm{TR}$ = 0.75 \\ \midrule
		Best & 0.927 $\pm$ 0.072 & 0.717 $\pm$ 0.276 & $\bm{0.513 \pm 0.102}$ & 0.849 $\pm$ 0.133 \\
		Last & 1.388 $\pm$ 0.494 & $\bm{0.884 \pm 0.280}$ & 0.943 $\pm$ 0.158 & 1.246 $\pm$ 0.209 \\
		\bottomrule
	\end{tabular}
\end{table} 

\begin{figure}[h!]
    \centering
    \includegraphics[width=0.75\textwidth]{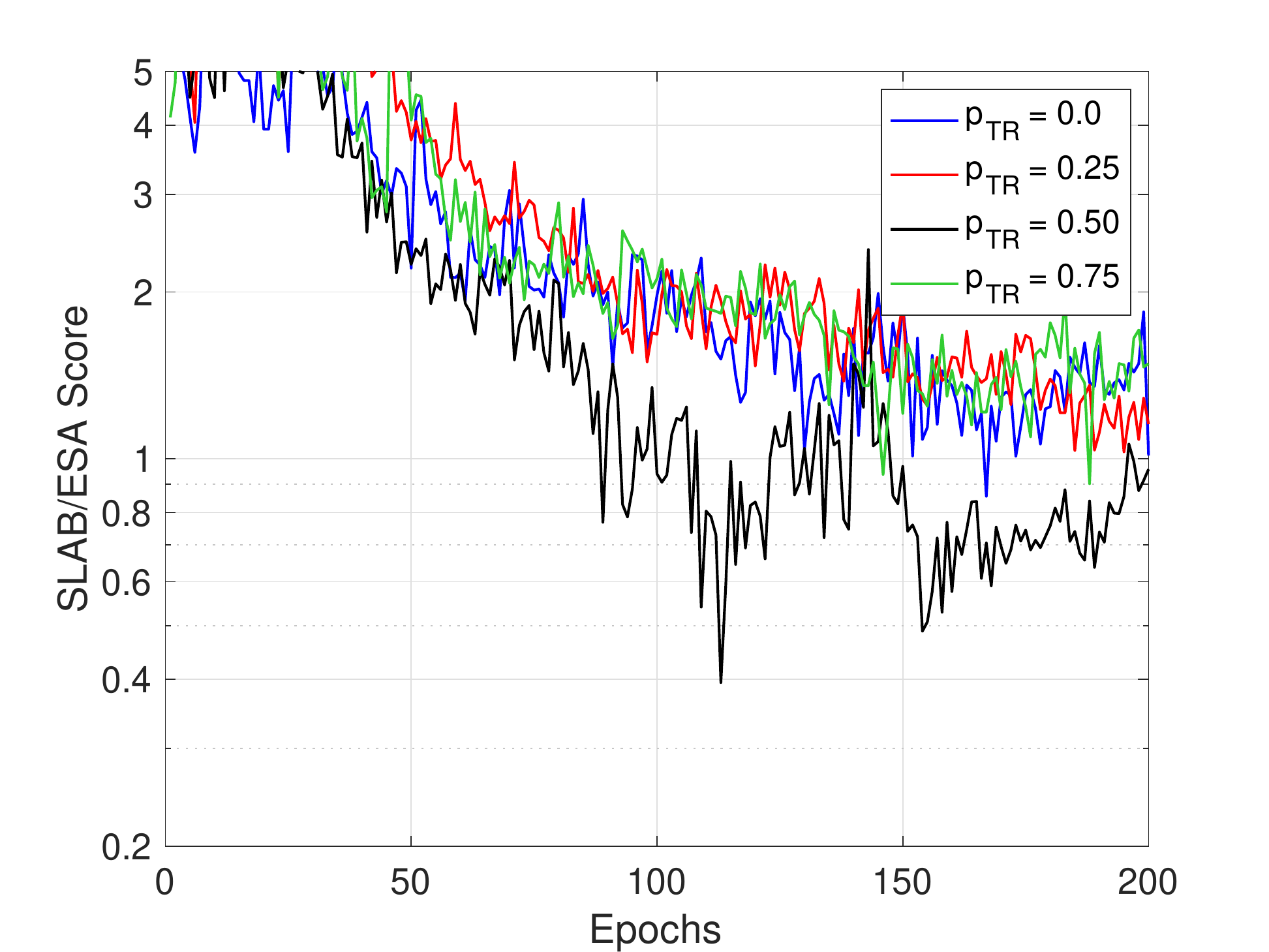}
    \caption{SLAB/ESA scores on PRISMA25 during training.}
    \label{fig:speedscore_tr_training}
\end{figure}

Table \ref{tab:performance_a25} reports the SLAB/ESA scores of the KRN-BB on PRISMA25 with varying $p_\textrm{TR}$. Specifically, the experiments are run three times with different random seeds to check the consistency in training behavior, and the averaged scores are reported using the network after the best-performing epoch (Best) and the last epoch (Last). First, according to the scores at the best-performing epochs, the KRN-BB with $p_\textrm{TR} = 0.5$ consistently achieves the lowest SLAB/ESA score compared to the network with other $p_\textrm{TR}$ or the network trained purely on synthetic PRISMA12K (i.e. $p_\textrm{TR} = 0$). However, the SLAB/ESA scores reported after the training is complete gives an impression that there is no visible improvement when the images from PRISMA12K-TR are introduced during training. The reason is that the network's performance on spaceborne images from PRISMA25 becomes very volatile as the training nears the end, as visible in Figure \ref{fig:speedscore_tr_training}. However, despite the volatility, it is obvious from Figure \ref{fig:speedscore_tr_training} that the training with $p_\textrm{TR} = 0.5$ outperforms the other study cases in general throughout the training.

\begin{figure}[t]
    \centering
    \includegraphics[width=\textwidth]{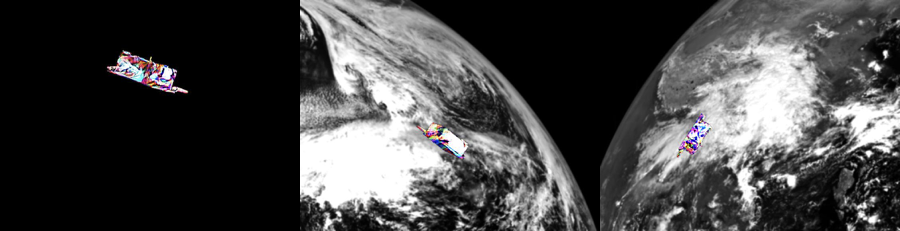}
    \caption{Examples of bad texture randomization.}
    \label{fig:prisma12k_bad_tr}
\end{figure}

One strong candidate for the cause of volatility on PRISMA25 is the fact that the texture randomization via NST inevitably disrupts the local shapes and geometric features of the spacecraft. For example,  Figure \ref{fig:prisma12k_bad_tr} visualizes three cases in which the asymmetric parts of the spacecraft, such as antennae and features on the spacecraft body, are rendered indistinguishable due to the NST pipeline. Then, it is not surprising that the evaluation on PRISMA25 becomes volatile, since the features that matter for regressing the bounding box corner locations are also unstable.

\begin{table}[t]
	\caption{Performance of both KRNs on PRISMA25 for varying $p_\textrm{TR}$. The average and standard deviation of SLAB/ESA score over 3 tests are reported for the best and last epochs. Bold face numbers represent the best performances.}
	\label{tab:performance_bb_kp}
	\centering
	\begin{tabular}{@{}ccccc@{}} \toprule
	    {} & \multicolumn{2}{c}{KRN-BB} & \multicolumn{2}{c}{KRN-SK} \\
	    {} & $p_\textrm{TR}$ = 0 & $p_\textrm{TR}$ = 0.5 & $p_\textrm{TR}$ = 0 & $p_\textrm{TR}$ = 0.5 \\ \midrule
		Best & 0.927 $\pm$ 0.072 & $\bm{0.513 \pm 0.102}$ & 1.117 $\pm$ 0.361 & 0.938 $\pm$ 0.062 \\
		Last & 1.388 $\pm$ 0.494 & $\bm{0.943 \pm 0.158}$ & 1.346 $\pm$ 0.252 & 1.048 $\pm$ 0.175 \\
		\bottomrule
	\end{tabular}
\end{table} 

Interestingly, the effect of losing the local shapes and geometric features is emphasized in Table \ref{tab:performance_bb_kp}, which compares the SLAB/ESA scores of KRN-BB and KRN-SK for $p_\textrm{TR}$ = 0 and 0.5. From the table, it is clear that when the images from PRISMA12K-TR are used during training, regressing the bounding box corners leads to consistently better performance than regressing the surface keypoints. At first glance, this is contrary to the trend in the state-of-the-art pose estimation methods based on CNNs that use surface keypoints due to their better connection to the object's features compared to the bounding box corners. However, it is likely that such connection leads to the degrading performance as the NST pipeline inevitably disrupts the local shapes and geometric features of the spacecraft to which the surface keypoints are tightly connected. Evidently, it is difficult even with human eyes to locate all the visible surface keypoints from the images of Figure \ref{fig:prisma12k_bad_tr}. Bounding box corners, on the other hand, may be instead better related to the global shape of the spacecraft which is not as damaged as the local shape by the NST pipeline.

\section{Conclusion}

This paper makes two contributions to the state-of-the-art in learning-based pose estimation of a noncooperative spacecraft using monocular vision. First, this work introduces a novel CNN architecture that merges the state-of-the-art detection and pose estimation pipelines with MobileNet architecture to enable fast and accurate monocular pose estimation. Specifically, the proposed CNN has mean rotation error of 3$^\circ$ and translation error under 25 cm when tested on the validation set, exceeding the performance of the state-of-the-art method\cite{Sharma2019}. Moreover, by cropping the original image using the detected RoI and designing the CNN to regress the 2D surface keypoints, the proposed architecture shows improved performance to images of both close and large inter-spacecraft separation compared to the state-of-the-art. Second, texture randomization is introduced as a vital step in training to enable the capability of CNN architectures in space missions. The keypoint regression network, when exposed to the texture-randomized images with 50\% probability during training, results in more accurate predictions from the spaceborne images it has not seen previously. However, the analysis reveals that the neural style transfer inadvertently disrupts the local features of the target, making keypoint regression more difficult and unstable. The superior performance of the bounding box corners compared to the surface keypoints suggests that the bounding box corners are less affected by the change of local features, but their performance on the spaceborne images is still volatile nonetheless. Future work should aim at developing the texture randomization technique with minimized effect on the object's local features.

There are few other challenges that still remain to be overcome. First of all, while the proposed CNN has real-time capability on desktop CPUs, the same on spacecraft hardware must be assessed in order to fully evaluate its applicability to an on-orbit servicing mission. Moreover, the current architecture assumes the knowledge of the spacecraft's shape and geometry. In reality, many mission scenarios that can benefit from accurate pose estimation, especially debris removal, cannot rely on the assumption of a known target model and instead must characterize its geometry autonomously during the encounter. Future research must address the problem of robust, efficient, and autonomous characterization of the target geometry and pose determination about an unknown spacecraft or debris.

\section{Acknowledgement}
The authors would like to thank the King Abdulaziz City for Science and Technology (KACST)
Center of Excellence for research in Aeronautics \& Astronautics (CEAA) at Stanford University for
sponsoring this work.



\bibliographystyle{AAS_publication}   
\bibliography{reference.bib}   

\end{document}